\pdfoutput=1
\documentclass[11pt]{article}

\usepackage[preprint]{acl}
\usepackage{enumitem}

\usepackage{times}
\usepackage{latexsym}

\usepackage{dblfloatfix}
\usepackage{tabularx}
\usepackage{booktabs}
\usepackage{array}
\newcolumntype{L}[1]{>{\raggedright\arraybackslash}p{#1}}
\newcolumntype{Y}{>{\raggedright\arraybackslash}X}

\newcommand{\var}[1]{\textsc{#1}} % small caps
\newcommand{\feat}[1]{\texttt{#1}}

\usepackage[T1]{fontenc}
\usepackage[utf8]{inputenc}

\usepackage{microtype}

\usepackage{inconsolata}

\usepackage{graphicx}
\usepackage{booktabs}
\title{\textsc{rumlem}: A Dictionary-Based Lemmatizer for Romansh}

\author{Dominic P. Fischer \quad Zachary Hopton \quad Jannis Vamvas \\
  Department of Computational Linguistics, University of Zurich\\
  \small
  \texttt{\{dominicphilipp.fischer, zacharywilliam.hopton, jannisnikos.vamvas\}@uzh.ch}}

\begin{document}
\maketitle
\begin{abstract}
Lemmatization -- the task of mapping an inflected word form to its dictionary form -- is a crucial component of many NLP applications. In this paper, we present \textsc{rumlem}, a lemmatizer that covers the five main varieties of Romansh as well as the supra-regional standard variety Rumantsch Grischun. It is based on comprehensive, community-driven morphological databases for Romansh, enabling \textsc{rumlem} to cover 77–84\% of the words in a typical Romansh text.
Since there is a dedicated database for each Romansh variety, an additional application of \textsc{rumlem} is variety-aware language classification.
Evaluation on 30'000 Romansh texts of varying lengths shows that \textsc{rumlem} correctly identifies the variety in 95\% of cases.
In addition, a proof of concept demonstrates the feasibility of Romansh vs. non-Romansh language classification based on the lemmatizer.

\vspace{0.5em}
\hspace{.5em}\includegraphics[width=1.25em,height=1.15em]{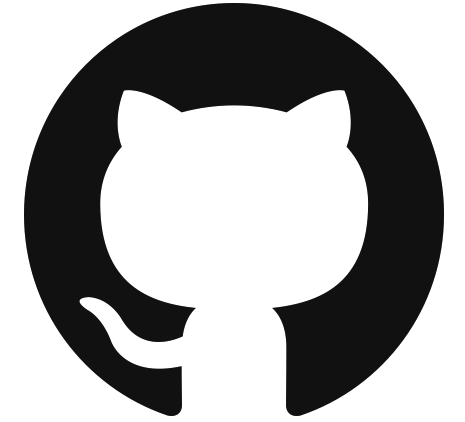}\hspace{.75em}
\parbox{\dimexpr\linewidth-7\fboxsep-7\fboxrule}{\url{https://github.com/ZurichNLP/rumlem}}
\vspace{-.5em}
\end{abstract}

\section{Introduction}
% INTRODUCTION
Romansh is a minority Romance language spoken by approximately 40'000--60'000 speakers in several Alpine valleys of Switzerland. It comprises five regional varieties, or \textit{idioms} (Sursilvan, Sutsilvan, Surmiran, Puter, and Vallader), as well as Rumantsch Grischun (RG), the supra-regional standard variety of Romansh. The varieties differ to such an extent that mutual intelligibility is often limited, highlighting the need for variety-specific NLP-tools. 

\begin{figure}[t]
    \centering
    \includegraphics[width=\columnwidth]{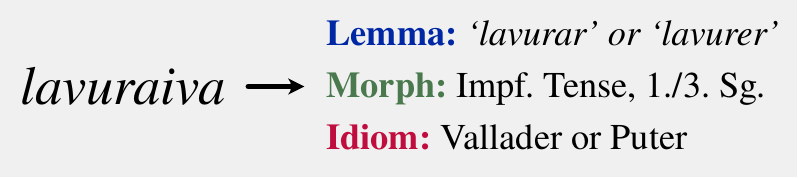}
\caption{The main functionalities of \textsc{rumlem}.}
    \label{fig:figure1}
\end{figure}

% MOTIVATION + CONTRIBUTION
The proposed Lemmatizer \textsc{rumlem}, given Romansh text, uses morphological databases to \textcolor[HTML]{0028A5}{\textbf{(1)}}~infer the possible lemmas of words forms, \textcolor[HTML]{397B49}{\textbf{(2)}}~identify morphological features of word forms, and \textcolor[HTML]{BF0D3E}{\textbf{(3)}}~identify the likely Romansh variety of the input (cf. Figure \ref{fig:figure1}). Together with \citet{model-2025-rumantsch-idiom}, our system is among the very few systems to reliably perform such a classification; with the additional benefit that it can be used to distinguish between Romansh and non-Romansh text. It thus provides a transparent complement to machine learning approaches for language identification and a core component to variety-aware Romansh NLP-tools. 

% METHOD
Our approach builds on existing, maintained, and community-driven dictionary data (cf. Table \ref{tab:pled_grond}), which we process into 725'005 unique word forms mappable to 178'467 lemmas (cf. Table \ref{tab:lem_data}).

% RESULTS
These data allow our lemmatizer to cover around 80\% of a typical Romansh text (cf. Section \ref{sec:coverage}) and to identify the variety correctly in 95\% of cases (cf. Table \ref{tab:Idiom_ID}). Language identification experiments (cf. Figure \ref{fig:50-300}) show that a threshold of ca. 0.6 (i.e., 60\% of words recognised as a particular Romansh variety) serves to distinguish Romansh texts from the most closely related Romance languages.

\section{Dictionary Resources for Romansh}

\begin{table*}[t]
\begin{tabularx}{\textwidth}{@{}X||r|rrrrr@{}} \toprule
         & Unique Entries & Single Words & DE Translations & POS Tags & Gender  & Infl. Verbs \\ \midrule
Sursilvan & 147,977         & 93,211       & 147,971             & 67,201   & 72,460  & 5,031       \\
Sutsilvan & 58,584          & 39,191       & 58,581              & 42,817   & 30,876  & 3,021       \\
Surmiran  & 74,986          & 44,365       & 74,986              & 39,224   & 36,956  & 2,947       \\
Puter     & 89,908          & 32,084       & 89,807              & 13,712   & 36,918  & 3,383      \\
Vallader  & 106,690         & 35,322       & 106,438             & 10,854   & 48,435  & 3,779      \\
RG        & 249,169         & 94,291       & 249,165             & 98,046   & 161,942 & 3,867    \\ 
\midrule
Total     & 727,314         &338,464       & 726,948             &271,854   & 387,587 & 22,028 \\
\bottomrule  
\end{tabularx}
\caption{Description of the \textit{Pledari Grond} dictionary for each Romansh idiom as well as Rumantsch Grischun. A single `|' means `thereof': Unique entries, \textit{thereof} X Single Words, German Translations, etc. \textit{Infl. Verbs} refers to the number of unique verbs for which inflected forms are provided.}
\label{tab:pled_grond}
\end{table*}

\subsection{Bilingual Dictionaries}
\textit{Pledari Grond}\footnote{\url{https://pledarigrond.ch/}}, the dictionary underlying \textsc{rumlem}, covers all six Romansh varieties, with translations provided in German (DE), as well as, in part, additional annotations (cf. Table \ref{tab:pled_grond}). Users may report potentially erroneous German–Romansh pairs and suggest alternative translations.

The dictionaries for Rumantsch Grischun, Surmiran, Sursilvan and Sutsilvan are openly licensed (© Lia Rumantscha 1980–2025). The Vallader and Puter dictionaries were kindly provided by Uniun dals Grischs for use solely as part of this lemmatizer (© Uniun dals Grischs. All rights reserved).

\subsection{Spellchecking}
\label{sec:Spellchecking}
Pledari Grond also provides a Romansh spell-checking system based on \textsc{Hunspell}\footnote{\url{https://hunspell.github.io/}}. With the focus lying primarily on orthographic conventions rather than on inflectional or derivational morphology across the different varieties, we do not use the spell-checker for inflectional processing. We restrict its use to providing a fallback vocabulary, together with the \textit{Mediomatix} corpus \cite{hopton2026mediomatix} and the Rumantsch Grischun newspaper \textit{La Quotidiana}\footnote{\url{https://huggingface.co/datasets/ZurichNLP/quotidiana}}. The lemmatizer uses said vocabulary to check non-lemmatizable words against a variety’s lexicon.

\section{Software Design}
\subsection{Preprocessing of Dictionary Data}
\label{sec:preprocessing_dict_data}

Lemma mappings constitute the central building block of the lemmatizer, making the transformation of the available dictionary data into this format a key step. We treat the morphologically rich and frequently annotated parts of speech (POS) nouns, verbs, and adjectives separately. Other POS (where present) or entries lacking POS tags were treated jointly. Where possible, we used conservative, rule-based heuristics to assign missing POS tags -- for example, treating entries whose German translation begins with a capital letter as nouns.

\begin{table*}[t]
\centering
\begin{tabular}{l||r|r|r|rrrr}
\toprule
 & Vocab & Mapped Forms & Lemmas & Noun & Adj & Verb & Other \\
\midrule
Sursilvan & 223,826 & 222,860 &     36,505    &    23,206    &    4,977    &   5,858     &    2,464        \\
Sutsilvan & 129,519 &  87,902 &     19,326    &    12,467    &    2,671    &   3,033     &    1,155      \\
Surmiran  & 149,078 &  84,481 &     22,838    &    15,145    &    3,107    &   3,204     &    1,382      \\
Puter     & 180,361 & 107,758 &     26,201    &    15,534    &    3,122    &   3,102     &    4,443      \\
Vallader  & 165,354 & 109,090 &     30,479    &    19,841    &    4,821    &   3,625     &    2,192      \\
RG  & 180,690 & 112,914 &     43,118    &    31,200    &    6,099    &   4,049     &    1,770      \\ 
\hline
Total   & 1,028,828 &     725,005   & 178,467    &    117,393    &    24,797   &     22,871  &    13,406
    \\ 
\bottomrule
\end{tabular}
\caption{\textsc{rumlem}'s data coverage, single words only. \textit{Mapped Forms} describes the amount of entries linked to a lemma. A single `|' means `thereof': Vocab, \textit{thereof} X Mapped Forms, \textit{thereof} X Lemmas, etc.}
\label{tab:lem_data}
\end{table*}

The entries in the Pledari Grond dictionaries exhibit a wide range of structural patterns, with e.g. approximately 200 distinct patterns each for nouns and adjectives. Pattern recognition distinguishes between single (w) and multiple words (w+), punctuation symbols, as well as special morphological tags (marked below as MT; e.g., m., f., sg., pl.). Two of the most frequent noun patterns and a less frequent one serve as illustration (\# occurrences):
\begin{itemize}[align=left, itemsep=0.0em, topsep=0.15em] 
    \item[\textbf{w}]  (208,000): armaziun; f; Bewaffnung
    \item[\textbf{w, w}]  (5067): admiratur, admiratura; m/f; Bewunderer(in)
    \item[\textbf{w (w, MT); w (w, MT)}] (93): arrestà (arrestats, pl); arrestada (arrestadas, pl); m/f; Gefangene
\end{itemize}

\begin{table*}[b]
\centering
\small
\renewcommand{\arraystretch}{1.15}

\begin{tabularx}{\textwidth}{@{}
  L{.15\textwidth}
  Y
  L{.12\textwidth}
@{}}
\toprule
\textbf{Variety} & \textbf{Form + features} & \textbf{Gloss} \\
\midrule
\var{rm-surmiran} & \feat{fomanto [PoS=ADJ; Gender=FEM; Number=SG]} & hungrig \\
\var{rm-surmiran} & \feat{fomantar [PoS=V; VerbForm=PTCP; Tense=PST; Gender=FEM; Number=SG]} & aushungern \\ \hline
\var{rm-vallader} & \feat{fomantà [PoS=ADJ; Gender=FEM; Number=SG]} & ausgehungert \\
\var{rm-vallader} & \feat{fomantà [PoS=ADJ; Gender=FEM; Number=SG]} & hungrig \\
\var{rm-vallader} & \feat{fomantada [PoS=N; Gender=FEM; Number=SG]} & Ausgehungerte \\
\var{rm-vallader} & \feat{fomantada [PoS=N; Gender=FEM; Number=SG]} & Hungrige \\
\var{rm-vallader} & \feat{fomantar [PoS=V; VerbForm=PTCP; Tense=PST; Gender=FEM; Number=SG]} & jn aushungern \\
\bottomrule
\end{tabularx}

\caption{\textsc{rumlem}'s output given the token `fomantada', with potential lemmas and morphological annotations of the form itself returned, as well as the German translation (variants).}
\label{tab:lem_options}
\end{table*}

Based on such recurring patterns, informed decisions could be made about how to process the data; for example, that a \textbf{w, w} entry such as the one above should yield two separate entries, \textit{admiratur; m; Bewunderer} and \textit{admiratura; f; Bewundererin}. To ensure clean and consistent processing, we automatically generated test skeletons for each distinct pattern occurring more than ten times in each of the four POS categories (N, V, ADJ and other), and manually annotated the corresponding gold-standard outputs. 200 such tests, covering 99.9\% of the input data, contribute to high data quality (cf. Appendix \ref{app:app_testcase}). Table \ref{tab:lem_data} presents the resulting data available to the lemmatizer.

\subsection{Variety Identification Process}
The lemmatizer takes a text and an optional variety. If none is given, it predicts the most likely variety based on the input text. More specifically, the text is tokenized using an adapted version\footnote{The adaptations consist of regex-based preprocessing and protected token patterns designed to correctly tokenize apostrophe-based contractions in each Romansh variety.} of the Italian Moses tokenizer~\cite{koehn-etal-2007-moses}; then, for each variety, the system counts lemmatizable tokens and tokens found in the variety-specific vocabulary and divides this count by the total number of tokens. 

\subsection{Lemmatization Process}
Consider the sentence \textit{La vuolp d’eira darcheu üna jada fomantada} (``The fox was once again hungry''). When processed by the lemmatizer, the variety is correctly identified as Vallader. Focusing on the token \textit{fomantada}, the lemmatizer returns the possible analyses shown in Table \ref{tab:lem_options}: assuming Vallader, the form may correspond to the feminine form of the adjective \textit{fomantà}, a feminine noun \textit{fomantada}, or the past participle of the verb \textit{fomantar}.

\section{Evaluation}

\subsection{Lemmatization Coverage}
\label{sec:coverage}
The data used for this task as well as for variety identification consisted of 3000--7000 texts for each variety, covering a range of input lengths. Shorter texts consist of validated speech transcripts from Romansh broadcasts by Radiotelevisiun Svizra Rumantscha (RTR). Texts longer than 300 tokens are taken from a set of children's stories called \textit{Babulins}, which exist in each Romansh variety. Note that the distributions are not even and differ between varieties; we report them in Appendix \ref{app:app_idiomID}.

We define lemmatization coverage as the percentage of word forms in a Romansh text for which the lemmatizer returns an analysis (i.e., excluding forms in the fallback-vocabulary). Removing the high-frequency punctuation symbols "$.,!?;:$", we find that \textsc{rumlem} lemmatizes around 80\% of all word forms, with variety-specific coverages ranging between 77\% and 84\%. We report detailed scores in Appendix \ref{app:app_coverage}. 

\subsection{Dictionary-based Variety Identification}
\label{sec:Idiom_ID}
 We also evaluated the performance of our lemmatizer in terms of variety classification accuracy. The results, summarized in Table \ref{tab:Idiom_ID}, show that \textsc{rumlem} accurately recognizes the variety of the vast majority of Romansh texts, especially longer texts. We note that the text genre might, in addition to text length, play a role in classification accuracy.

\begin{table}[h]
\centering
\small
\renewcommand{\arraystretch}{1.15}
{\setlength{\tabcolsep}{4pt}
\begin{tabular}{@{}lrrrrr|r@{}}
\toprule
\textbf{Length} & \textbf{2--10} & \textbf{10--50} & \textbf{50--300} & \textbf{300--800} & \textbf{800+} & \textbf{All} \\
\midrule
Sursilvan & 0.85 & 0.85 & 0.87 & 1.00 & 1.00 & 0.86 \\
Sutsilvan & 1.0  & 0.99 & 1.0  & 1.00 & 1.00 & 1.00 \\
Surmiran  & 0.92 & 0.94 & 0.99 & 1.00 & 1.00 & 0.95 \\
Puter     & 0.97 & 0.98 & 0.99 & 1.00 & 1.00 & 0.98 \\
Vallader  & 0.94 & 0.91 & 0.93 & 1.00 & 1.00 & 0.92 \\
RG        & 0.89 & 1.00 & 1.00 & 1.00 & 1.00 & 1.00 \\
\hline
All       & 0.94 & 0.94 & 0.95 & 1.00 & 1.00 \\  
\bottomrule
\end{tabular}}
\caption{Classification accuracy by Rumantsch variety across length buckets (number of tokens).}
\label{tab:Idiom_ID}
\end{table}

These scores are comparable to what \citet{model-2025-rumantsch-idiom} reports on balanced in-domain data, as well as unbalanced in-domain data with longer samples (avg. ca. 530 tokens). On shorter unbalanced in-domain data (avg. ca. 85 tokens) and out-of-domain data, their SVM classifier struggles, with F1 scores dropping to ca. 0.8 and 0.7, respectively.

\subsection{Dictionary-based Language Identification}
\begin{figure}[h]
    \centering
    \includegraphics[width=\columnwidth]{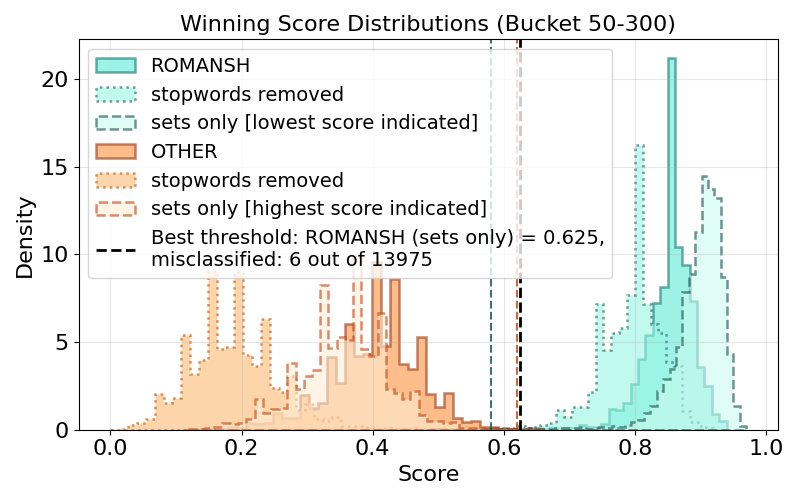}
\caption{Distributions of Romansh (turquoise) and other Romance languages (rust) according to the highest Romansh variety score assigned to each text sample.}
    \label{fig:50-300}
\end{figure}

We selected about 5000 texts from Fineweb\footnote{The data is made up of webpages crawled by CommonCrawl between 2013 and 2024.} in Romansh itself as well as the four Romance languages French, Italian, Catalan and Romanian, as these languages are typologically close to Romansh and therefore most likely to exhibit overlapping dictionary forms (cf. Table \ref{app:app_LID}). We record the ``winning'' scores, i.e., the highest score assigned to a text across the Romansh varieties. Figure~\ref{fig:50-300} shows the Romansh score distributions in turquoise and the non-Romansh ones in rust color for three different setups: as-is, using the sets of words, and removing Romance-language (FR, IT, CA, RO) stopwords.

Figure \ref{fig:50-300} and App. \ref{app:app_LIDthresholds} show that, using the sets of words, a separating threshold can be found for all three tested buckets. Perfect separation was achieved apart from bucket 50--300; however, each of the misclassified samples was highly noisy, containing a mix of languages (cf. App. \ref{app:app_LIDmisclassified}). Further manual inspection revealed the presence of many similar samples on the lower end of the Romansh distribution; the ideal threshold may thus lie higher.

These results suggest that a straightforward Romansh language identification system could be built using \textsc{rumlem}. Since the distributions are well-separated, a small validation set would suffice to determine the optimal classification threshold. 

\section{Related Work}

\paragraph{NLP for Romansh}
Romansh and its varieties are not yet covered by popular NLP tools and resources such as SpaCy, Universal Dependencies~\cite{de-marneffe-etal-2021-universal}, and UniMorph~\cite{batsuren-etal-2022-unimorph}, motivating the development of dedicated tools.
Recent years have nonetheless seen progress in other areas of Romansh NLP, including contextualized token embeddings and named entity recognition~\cite{vamvas-etal-2023-swissbert}, word alignment~\cite{dolev-2023-mbert}, and machine-learning-based variety identification~\cite{model-2025-rumantsch-idiom}.
This paper extends this line of work by providing a dictionary-based system for (context-agnostic) lemmatization and morphosyntactic analysis that, as we show, can also serve as a basis for variety and language identification.

\paragraph{Dictionary-based Lemmatization}
While neural approaches to lemmatization and morphosyntactic analysis can take into account the context of word forms and even generalize to unseen forms~\cite{straka-etal-2016-udpipe,mccarthy-etal-2019-sigmorphon,qi-etal-2020-stanza}, they require supervised data, typically in the form of treebanks.
In the absence of such treebanks, dictionary-based lemmatization is a viable alternative when a comprehensive dictionary is available for a language.
Dictionary-based lemmatizers have been proposed for, among others, German (based on Wiktionary;~\citealp{liebeck-conrad-2015-iwnlp}), Middle English~\cite{karimov2016hybrid}, Latin~\cite{passarotti-etal-2017-lemlat}, and Somali~\cite{mohamed2023lexicon}.
%A drawback of such approaches is a lack of context-sensitivity and limitation to a fixed lexicon; moreover, the feature set is constrained by the information available in the underlying dictionary.
Our system builds on six large-scale, highly consistent dictionaries for the Romansh varieties that include inflection tables and German translations, enabling relatively high word coverage and a rich feature set.

\section{Conclusion}
We presented \textsc{rumlem}, a dictionary-based lemmatizer covering all six Romansh varieties. Beyond lemmatizing around 80\% of a given Romansh text, \textsc{rumlem} reliably identifies Romansh varieties -- averaging 95\% accuracy across varieties and text lengths -- and can be used to distinguish Romansh even from its most closely related Romance languages. \textsc{rumlem}'s transparent design makes it a useful complement to machine learning approaches for Romansh NLP.

\section*{Limitations}
\textsc{rumlem}'s performance is inherently bounded by its dictionary coverage and quality: words absent from Pledari Grond cannot be lemmatized, and annotation errors will be propagated as-is. Since the lemmatizer cannot account for context, ambiguous forms may receive multiple analyses without disambiguation. 
Future work could explore the use of statistical or neural approaches to make the lemmatizer more context-aware.

Finally, we note that while the software of \textsc{rumlem} itself is released open-source including the postprocessed dictionary data, the Vallader and Puter dictionaries are released without an open-source license, and use of these dictionaries for research beyond \textsc{rumlem} will require written permission from the copyright holders.

\section*{Acknowledgments}
% List looks good to Ignacio; no preferences for exact phrasing
We thank the Swiss Federal Office of Culture, the Lia Rumantscha, and the Uniun dals Grischs for their support, the participants of the digidi 2025 workshop for fruitful discussions, and Sina Ahmadi for helpful feedback.

% Bibliography entries for the entire Anthology, followed by custom entries
%\bibliography{anthology,custom}
% Custom bibliography entries only
\bibliography{latex/custom}

%\appendix
\appendix

\section{Preprocessing Tests}
\label{app:app_testcase}

The example below illustrates a preprocessing test case. The first line shows the raw dictionary data, while the lines following '$>>>$' show the validated format -- i.e., how the data is processed before being fed into the lemmatizer.”

\begin{verbatim}
'antalg(iant)evel, antalg(iant)evla'; adj
>>> antalgevel:
        antalgevel; ADJ;MASC;SG
        antalgevla; ADJ;FEM;SG
    antalgiantevel:
        antalgiantevel: ADJ;MASC;SG
        antalgiantevla: ADJ;FEM;SG
\end{verbatim}  

\section{Lemmatization Coverage}
\label{app:app_coverage}

Table \ref{tab:coverage} shows coverage values across texts of varying lengths and varieties. The text data used are the same as in the variety identification experiments -- shorter texts come from Fineweb, longer ones from Babulins (cf. Section \ref{sec:coverage}). Overall, coverage does not vary significantly with text length or genre, except for fragmentary texts, where individual missing word forms have a greater impact.

\begin{table}[h]
\centering
\small
\renewcommand{\arraystretch}{1.15}
{\setlength{\tabcolsep}{4pt}
\begin{tabular}{@{}lrrrrr|r@{}}
\toprule
\textbf{Length} & \textbf{2--10} & \textbf{10--50} & \textbf{50--300} & \textbf{300--800} & \textbf{800+} & \textbf{All} \\
\midrule
Surs. & 0.76 & 0.81 & 0.84 & 0.83 & 0.78 & 0.84 \\
Suts. & 0.71 & 0.77 & 0.77 & 0.73 & 0.73 & 0.77 \\
Surm. & 0.73 & 0.80 & 0.84 & 0.82 & 0.79 & 0.82 \\
Puter & 0.79 & 0.83 & 0.84 & 0.84 & 0.81 & 0.84 \\
Vall. & 0.66 & 0.77 & 0.80 & 0.83 & 0.81 & 0.80 \\
RG    & 0.93 & 0.79 & 0.79 & 0.80 & 0.78 & 0.79 \\
\hline
All   & 0.75 & 0.80 & 0.82 & 0.81 & 0.79 &  \\
\bottomrule
\end{tabular}}
\caption{Coverage ratios by text length and variety. Values for each variety and bucket are averaged across samples. ‘All’ shows the total number of lemmatizable tokens divided by the total tokens in the bucket/variety, not an average.}
\label{tab:coverage}
\end{table}

Of the approximately 40,000 missing (i.e., un-lemmatized) tokens across all varieties, around 11,000 are proper nouns or German nouns. Another 4,000 consist of tokens containing numbers, special tokens, or characters such as dashes and hyphens. Notable cases include contractions that are absent from the Pledari Grond dictionaries. The remaining 25,000 tokens appear to be Romansh words that are not yet included in the lemmatizer.

\section{Variety ID}
\label{app:app_idiomID}

\subsection{Samples per Variety and Bucket}

Table \ref{tab:app_idiomid} shows the number of samples per Romansh variety and bucket. The same samples were used for both coverage evaluation as well as variety identification.

\begin{table}[h]
\centering
\small
\renewcommand{\arraystretch}{1.15}
{\setlength{\tabcolsep}{4pt}
\begin{tabular}{@{}lrrrrr|r@{}}
\toprule
\textbf{Length} & \textbf{2--10} & \textbf{10--50} & \textbf{50--300} & \textbf{300--800} & \textbf{800+} & \textbf{Tot} \\
\midrule
Surs. & 68 & 2647 & 4173 & 7 & 5 & 6900 \\
Suts. & 6 & 1190 & 1795 & 5 & 7 & 3003 \\
Surm. & 1113 & 3751 & 2204 & 6 & 6 & 7080 \\
Puter & 660 & 2783 & 2468 & 5 & 7 & 5923 \\
Vall. & 277 & 2209 & 3202 & 5 & 7 & 5700 \\
RG & 9 & 1218 & 3052 & 6 & 6 & 4291 \\
\hline
Tot & 2133 & 13798 & 16894 & 34 & 38 & 29897 \\
\bottomrule
\end{tabular}}
\caption{\#samples by Romansh variety across buckets.}
\label{tab:app_idiomid}
\end{table}

\section{Language ID}
\label{app:app_LID}

\subsection{Samples per Variety and Bucket}
\label{app:app_LIDsamples}

Table \ref{tab:app_Lid} shows the number of samples per Romance variety and bucket, used for Romansh vs. non-Romansh identification.

\begin{table}[h]
\centering
\small
\renewcommand{\arraystretch}{1.15}
{\setlength{\tabcolsep}{5pt}
\begin{tabular}{@{}lrrr|r@{}}
\toprule
\textbf{Length} & \textbf{50--300} & \textbf{300--800} & \textbf{800--2000} & \textbf{Tot} \\
\midrule
French & 2517 & 1551 & 693 & 4761 \\
Italian & 2493 & 1620 & 671 & 4784 \\
Romanian & 2128 & 1513 & 922 & 4563 \\
Catalan & 2595 & 1575 & 593 & 4763 \\
Romansh & 4242 & 661 & 75 & 4978 \\
\hline
Tot & 13975 & 6920 & 2954 & 23849 \\
\bottomrule
\end{tabular}
}
\caption{\#samples by Romance language across buckets.}
\label{tab:app_Lid}
\end{table}

\subsection{Separating Thresholds per Bucket}
\label{app:app_LIDthresholds}

Note that the best threshold is defined as, primarily, the one that best separates the data, and, secondarily, the one with the widest margin of separation. In both buckets 300--800 and 800--2000, all methods resulted in perfect separation. However, using the sets of words provided the widest margin, indicating that using sets results in the best separation across the different buckets.

\begin{figure}[h]
    \centering
    \includegraphics[width=\columnwidth]{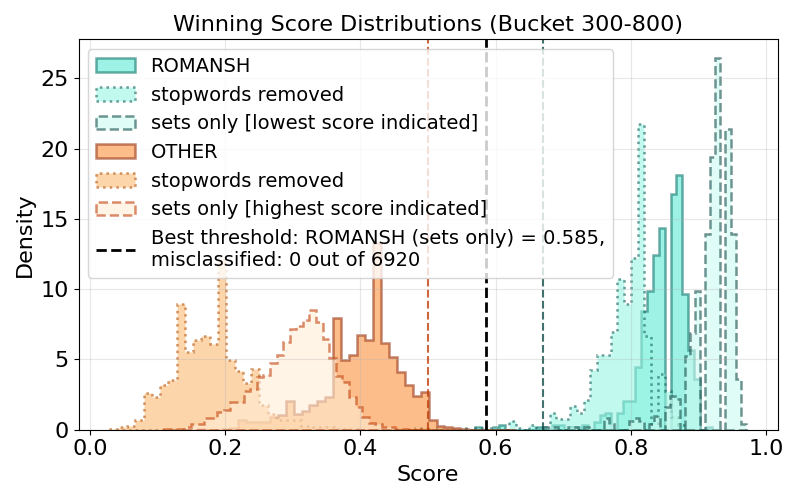}
\caption{Romansh (turqoise) and other Romance languages (rust) winning variety score distributions in the bucket with token length 300--800.}
    \label{fig:300-800}
\end{figure}

\begin{figure}[h]
    \centering
    \includegraphics[width=\columnwidth]{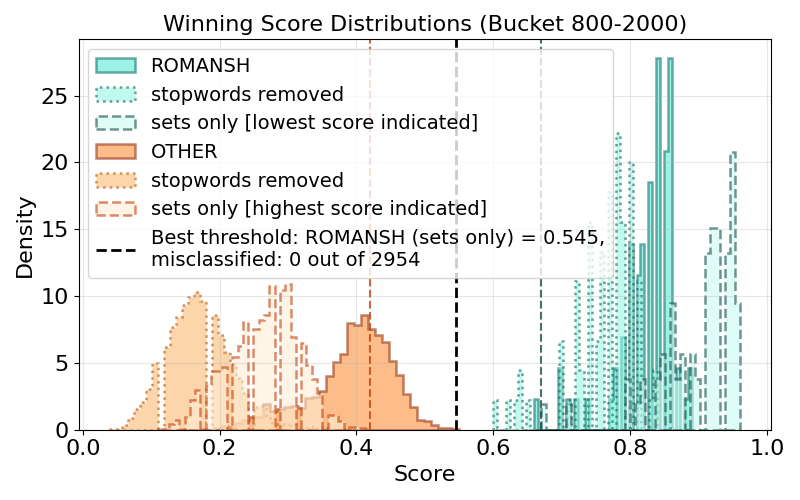}
\caption{Romansh (turqoise) and other Romance languages (rust) winning variety score distributions in the bucket with token length  800--2000.}
    \label{fig:800-2000}
\end{figure}

This was confirmed when we repeated the experiment with different Romansh data, namely the data from the variety classification task. Here too, reducing the texts to their sets of words before being processed by the lemmatizer produced the best threshold in all three token budget settings.

For all experiment settings, we also calculated the average score distribution across varieties instead of the winning score, which resulted in slightly more misclassifications and less interpretable thresholds. This is due to the non-trivial differences between the Romansh varieties, which should be treated separately instead of being conflated.

Finally, note that the ideal threshold decreasing with text length is due to longer texts widening the gap between the Romansh and non-Romansh text, whereby the threshold is placed in the middle.

\subsection{Analysis of Misclassified Samples}
\label{app:app_LIDmisclassified}

The six misclassified data samples from Figure \ref{fig:50-300} were all highly noisy. Four out of the six contained parallel translations (Romansh in bold, Italian in italic):

\begin{itemize}
    \item ``[..] \textbf{II cussagl da scoula as cumpuona da duos commembers e dal suprastant dal decasteri.} [...] Der Schulrat setzt sich aus zwei Mitgliedern zusammen sowie dem für diesen Bereich zuständigen Gemeinderatsmitglied. [...]''
%``Cussagl da scoula / Schulrat II cussagl da scoula as cumpuona da duos commembers e dal suprastant dal decasteri. Quist surpiglia il presidi, scha'l cussagl da scoula nu's constituischa svess. Al cussagl da scoula suprasto l'administraziun scolastica. El tscherna e licenzchescha ils magisters e las magistras, decretescha ün uorden da scoula chi stu gnir appruvo da la radunanza cumünela e ho l'incumbenza da pisserer pel bun uorden e la nettaschia aint ils locals da scoula. Las incumbenzas e'ls dovairs as drizzan tenor l'uorden da scoula. Der Schulrat setzt sich aus zwei Mitgliedern zusammen sowie dem für diesen Bereich zuständigen Gemeinderatsmitglied.. Dem Schulrat obliegt die Verwaltung der Schule. Er wählt und entlässt die Lehrpersonen, Der Schulrat ist beauftragt, für gute Ordnung und Reinlichkeit in den Schullokalen zu sorgen. Die Aufträge und Pflichten richten sich nach dem Schulgesetz.''
    \item ``Herzlich Willkommen Wir begrüssen Sie herzlich auf der Seite unserer Kirchgemeinde und danken Ihnen für Ihr Interesse an unserem Pfarreileben. [...] \textbf{Cordial beinvegni silla pagina dalla pleiv catolica Sevgein/Castrisch/Riein. Nus engraziein a Vus che Vus s’interesseis per nossa pleiv.} [...]''
%``Herzlich Willkommen Wir begrüssen Sie herzlich auf der Seite unserer Kirchgemeinde und danken Ihnen für Ihr Interesse an unserem Pfarreileben. Wir hoffen, dass diese Seiten dazu beitragen, dass Sie uns näher kennen lernen und so auch unser grosses Projekt, den Neubau der Orgel kennen lernen.  Gerne stehen wir Ihnen für weitere Informationen zur Verfügung. Nehmen Sie mit uns Kontakt auf – wir freuen uns darauf. Cordial beinvegni silla pagina dalla pleiv catolica Sevgein/Castrisch/Riein. Nus engraziein a Vus che Vus s’interesseis per nossa pleiv. Nus sperein che quellas paginas seigien in agid per emprender d’enconuscher empau nus. Vus haveis era la pusseivladad da dar in’egliada en nies grond project, l’installaziun dalla nov’orgla en la baselgia Sogn Tumasch. Bugen stein a Vossa disposiziun per ulteriuras informaziuns. Ei legra nus, sche Vus semetteis en contact cun nus. Katholische Kirchgemeinde Sevgein/Castrisch/Riein Pleiv catolica Sevgein/Castrisch/Riein''
    \item ``\textbf{Gorbatschow und Freund Sbalzs classics sün las cordas L’interpret da Balalaika straordinari da nos temp es il virtuos Prof. Andreij Gorbatschow chi viva a Moskau.} [...] Klassische Saitensprünge Der herausragende Balalaika-Interpret unserer Zeit ist der in Moskau lebende Star-Virtuose Prof. Andreij Gorbatschow. [...]''
    \item ``\textit{Un cordiale benvenuto} – Herzlich willkommen - \textbf{Cordial bainvegni} \textit{Associazione Spitex dei Grigioni Siamo l’associazione mantello delle 19 organizzazioni Spitex che operano nel Canton dei Grigioni.} [...] Spitex Verband Graubünden Wir sind der Dachverband der 19 im Kanton Graubünden tätigen Spitex-Organisationen. [...] \textbf{Federaziun grischuna da spitex Nus essan l'uniun tetgala da las 19 organisaziuns da spitex activas en il chantun Grischun.} [...]''
%``Un cordiale benvenuto – Herzlich willkommen - Cordial bainvegni Associazione Spitex dei Grigioni Siamo l’associazione mantello delle 19 organizzazioni Spitex che operano nel Canton dei Grigioni. In veste di associazione di settore e dei datori di lavoro tuteliamo gli interessi delle Spitex senza scopo di lucro grigionesi in campo sanitario e sociale nonché in ambito politico. Spitex Verband Graubünden Wir sind der Dachverband der 19 im Kanton Graubünden tätigen Spitex-Organisationen. Als Arbeitgeber- und Fachverband vertreten wir die Interessen der Bündner NPO-Spitex im Gesundheits- und Sozialwesen sowie im politischen Umfeld. Federaziun grischuna da spitex Nus essan l'uniun tetgala da las 19 organisaziuns da spitex activas en il chantun Grischun. Sco federaziun dals patruns ed associaziun professiunala represchentain nus ils interess da las spitexs grischunas, tut organisaziuns da non-profit, en dumondas da la sanadad, dals fatgs socials sco era da la politica.''
\end{itemize}

The remaining two contained what looked like web-scraping artifacts in German:
\begin{itemize}
    \item ``Foto aus dem Akt-Channel Teilnahme am Forum Fotos verkaufen Mehr Foto-Ordner anlegen? Mehr Fotos speichern? [...] \textbf{Igl october vargau havein nus era visitau quei marcau ed jeu muossel in maletg ord il casti da Schönbrunn. Amicabels salids giu da Glion Glieci}''
%``Foto aus dem Akt-ChannelTeilnahme am ForumFotos verkaufenMehr Foto-Ordner anlegen?Mehr Fotos speichern?Am Galerie-Voting teilnehmenProfilbesucher anzeigenMehr Fotos veröffentlichen?Zugang in allen Sprachversionen?WerbefreiheitPremium-Nutzer können schon ab 4 € im Monat alle Funktionen der fotocommunity nutzen.Bien gi Werner in bellezia maletg da Vienna muossas Ti cheu, en ina perspectiva nundetga. Igl october vargau havein nus era visitau quei marcau ed jeu muossel in maletg ord il casti da Schönbrunn. Amicabels salids giu da Glion Glieci''
    \item ``\textbf{Cla Rauch ha orientà davart l’Archiv Cultural d’Engiadina Bassa (fotografia: Benedict Stecher).} [...] Haben Sie noch kein Konto? Registrieren Sie sich hier [...]''
%``Cla Rauch ha orientà davart l’Archiv Cultural d’Engiadina Bassa (fotografia: Benedict Stecher). Ils seniors da Sent s’han inscuntrats illa Chasa Misoch per gnir infuormats davart l’Archiv Cultural d’Engiadina Bassa (ACEB) a S-chadatsch. Cla Rauch dal ACEB ha infuormà davart lur lavur per salvar perdüttas da temps passats. Haben Sie noch kein Konto? Registrieren Sie sich hier Einige Beiträge auf engadinerpost.ch können kommentiert werden. Bitte beachten Sie dazu folgende Punkte:''
\end{itemize}

\end{document}